\documentclass[11pt,a4paper]{article}
\usepackage[hyperref]{naaclhlt2019}
\usepackage{times}
\usepackage{latexsym}

\usepackage{url}

\aclfinalcopy 

\usepackage{booktabs}       
\usepackage{amsfonts}       
\usepackage{nicefrac}       
\usepackage{microtype}      

\usepackage{xspace}
\usepackage{xcolor}
\usepackage{graphicx}
\usepackage{wrapfig}
\usepackage{tabulary}
\usepackage{multirow}
\usepackage{capt-of}
\usepackage{makecell}
\usepackage{enumitem}
\usepackage{amsmath}
\usepackage{mathrsfs}
\usepackage[capitalize]{cleveref}

\let\vec\mathbf

\newcommand{\BiLSTM}{\text{BiLSTM}}
\newcommand{\softmax}{\text{softmax}}

\title{Bidirectional Attentive Memory Networks for Question Answering
over Knowledge Bases}

\author{Yu Chen \\
   Rensselaer Polytechnic Institute \\
  {\tt cheny39@rpi.edu} \\\And
  Lingfei Wu \\
  IBM Research \\
  {\tt wuli@us.ibm.com} \\\And
    Mohammed J. Zaki \\
  Rensselaer Polytechnic Institute \\
  {\tt zaki@cs.rpi.edu}}

\date{}

\begin{document}
\maketitle
\begin{abstract}
   When answering natural language questions over knowledge bases (KBs),
   different question components and KB aspects play different roles.
   However, most existing embedding-based methods for knowledge base question answering (KBQA) ignore the subtle inter-relationships
	between the question and the KB (e.g., entity types, relation paths and context).
    In this work, we propose to directly model the two-way flow of interactions between the
    questions and the KB via a novel Bidirectional Attentive Memory Network, called BAMnet.
    Requiring no external resources and only very few hand-crafted features, on the WebQuestions
    benchmark, our method significantly outperforms existing information-retrieval 
    based methods, and remains competitive with (hand-crafted) semantic parsing based methods. Also, since we use attention mechanisms, our method offers better interpretability compared to other baselines.  

\end{abstract}

\section{Introduction}

With the rapid growth in large-scale knowledge bases (KBs) such as
DBPedia \cite{auer2007dbpedia} and FreeBase \cite{freebase:datadumps},
knowledge base question answering (KBQA) has drawn increasing attention
over the past few years. {\em Given questions in natural language (NL), the goal
of KBQA is to automatically find answers from the underlying KB}, which
provides a more natural and intuitive way to access the vast underlying
knowledge resources.

One of the most prominent challenges of KBQA is the lexical gap. For instance,
the same question can be expressed in various ways in NL while a KB
usually has a canonical lexicon. It is therefore nontrivial to map an NL
question to a structured KB. The approaches proposed to tackle
the KBQA task can be roughly categorized into two groups: semantic
parsing (SP) and information retrieval (IR) approaches.
SP-based approaches address the problem by constructing a semantic
parser that converts NL questions into intermediate logic forms, which
can be executed against a KB. Traditional semantic
parsers \cite{wong2007learning}
require annotated logical forms as supervision,
and are limited to narrow domains with a small number of logical
predicates. Recent efforts overcome these limitations via the
construction of hand-crafted rules or features 
\cite{abujabal2017automated,hu2018answering}
schema matching \cite{cai2013large}, and using weak
supervision from external resources \cite{krishnamurthy2012weakly}. 

Unlike SP-based approaches that usually assume a pre-defined
set of lexical triggers or rules, which limit their domains and
scalability, IR-based approaches directly retrieve answers from the KB
in light of the information conveyed in the questions. These IR-based approaches usually do
not require hand-made rules and can therefore scale better to large and complex KBs.
Recently, deep neural networks have been shown to produce strong
results on many NLP tasks.
In the field of KBQA, under the
umbrella of IR-based approaches, many embedding-based methods
\cite{bordes2014open,hao2017end}
have been proposed
and have shown promising results. These methods adopt various ways to
encode questions and KB subgraphs into a common embedding space and
directly match them in that space, and can be typically trained in an
end-to-end manner.


Compared to existing embedding-based methods that encode questions and KB subgraphs independently, we introduce a novel
\textbf{B}idirectional \textbf{A}ttentive \textbf{M}emory \textbf{net}work, called \textbf{BAMnet}
that captures the mutual interactions between questions and the
underlying KB, 
which is stored in a content-addressable memory. We assume that the world knowledge (i.e.,
the KB) is helpful for better understanding the questions. Similarly, the questions themselves can help us focus on important KB aspects. 
To this end, we design a {\em two-layered bidirectional attention network}.
The {\em primary attention network} is intended to focus on important parts of a 
question in light of the KB and important KB aspects in light of the question.
Built on top of that, the {\em secondary attention network} 
is intended to enhance the question and KB representations 
by further exploiting the two-way attention.
Through this idea of {\em hierarchical two-way attention}, we are able
to distill the information that is the most relevant to answering the
questions on both sides of the question and KB.

 We highlight the contributions of this paper as follows:
 1) we propose a novel bidirectional attentive memory network for the task of KBQA which is
     intended to directly model the two-way interactions between questions and the
     KB;
 2) by design, our method offers good interpretability thanks to the
     attention mechanisms;
 3) on the WebQuestions benchmark, our method significantly outperforms previous information-retrieval 
    based methods while remaining competitive with (hand-crafted) semantic parsing based methods.

\section{Related work}
Two broad classes of SP-based and IR-based approaches have been proposed for KBQA.
The former attempts to convert NL questions to logic forms. 
Recent work focused on approaches based on weak supervision from either external resources
\cite{krishnamurthy2012weakly,berant2013semantic,yao2014information,hu2018answering,yih2015semantic,yavuz2016improving}, schema
matching \cite{cai2013large}, or using hand-crafted rules and
features
\cite{unger2012template,berant2013semantic,berant2015imitation,reddy2016transforming,bao2016constraint,abujabal2017automated,hu2018answering,bast2015more,yih2015semantic}.
A thread of research has been explored to generate semantic query graphs from NL questions 
such as using coarse alignment between phrases and predicates \cite{berant2013semantic},
searching partial logical forms via an agenda-based 
strategy \cite{berant2015imitation}, pushing down the disambiguation step into the query evaluation stage \cite{hu2018answering}, or exploiting rich syntactic information in NL questions \cite{xu2018graph2seq,xu2018exploiting}. 
Notably, another thread of SP-based approaches try to exploit IR-based techniques \cite{yao2014information,bast2015more,yang2014joint,yih2015semantic,bao2016constraint,yavuz2016improving,liang2016neural} 
by computing the similarity of two sequences as features, 
leveraging a neural network-based answer type prediction model,
or training end-to-end neural symbolic machine via REINFORCE~\cite{williams1992simple}.
However, most SP-based approaches more or less rely on hand-crafted rules or features, 
which limits their scalability and transferability.

The other line of work (the IR-based) has focused on mapping answers and questions into the same embedding space, where one could query any KB independent of its schema without requiring any grammar or lexicon. 
\citet{bordes2014open} were the first to apply an
embedding-based approach for KBQA. 
Later, \citet{bordes2014question} proposed the idea of subgraph embedding, which encodes more information (e.g., answer path and context) about the candidate answer. In follow-up work \cite{bordes2015large,jain2016question}, 
memory networks \cite{weston2014memory} were used to store candidates, and could be accessed iteratively to mimic multi-hop reasoning. Unlike the above methods that mainly use a
bag-of-words (BOW) representation to encode questions and KB resources,
\cite{dong2015question,hao2017end} apply more advanced network modules
(e.g., CNNs and LSTMs) to encode questions. Hybrid methods have also
been proposed \cite{feng2016hybrid,xu2016question,das2017question}, which achieve
improved results by leveraging additional knowledge sources such as
free text. While most embedding-based approaches encode
questions and answers independently,
\cite{hao2017end} proposed a cross-attention mechanism to
encode questions according to various candidate answer aspects. Differently, in this work, our method goes one step further by modeling the bidirectional
interactions between questions and a KB. 

The idea of bidirectional attention proposed in this work is similar to those applied in machine reading comprehension
\cite{wang2016machine,seo2016bidirectional,xiong2016dynamic}.
 However, these previous works focus on capturing the interactions between two bodies of text, in this work, we focus on modeling the interactions between one body of text and a KB.

\section{A modular bidirectional attentive memory network for KBQA}

\begin{figure*}[!tbh]
  \centering
    \includegraphics[keepaspectratio=true,scale=0.32]{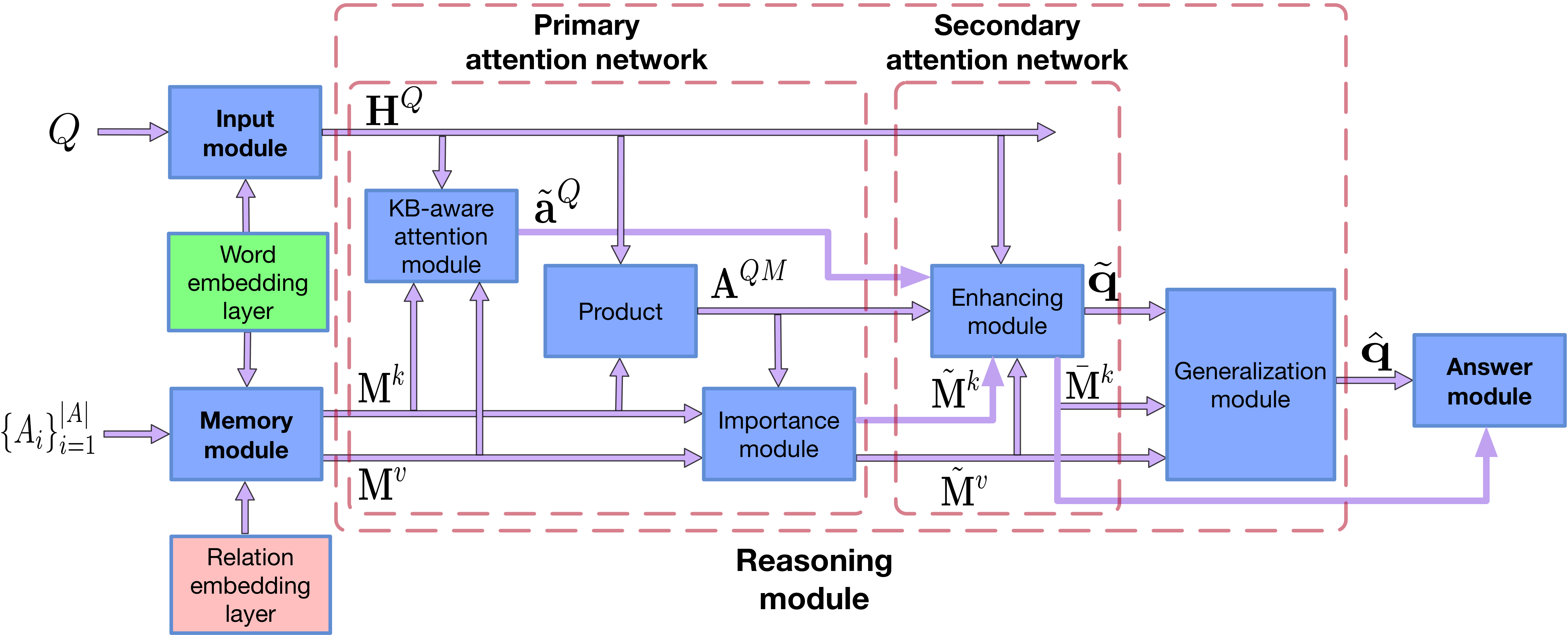}
  \caption{Overall architecture of the BAMnet model.}
  \label{fig:over_arch}
\end{figure*}

Given an NL question, the goal is to fetch answers from the underlying KB.
Our proposed BAMnet model consists of four components which are the {\em input module}, 
{\em memory module}, {\em reasoning module} and {\em answer module}, 
as shown in \cref{fig:over_arch}. 

\subsection{Input module}
\label{sec:input_module}

An input NL question $Q = \{q_i\}_{i=1}^{|Q|}$ 
is represented as a sequence of word embeddings ($q_i$) 
by applying a word embedding layer. We
then use a bidirectional LSTM \cite{hochreiter1997long}
to encode the question as $\vec{H}^Q$ (in $\mathbb{R}^{d \times |Q|}$) which 
is the sequence of hidden states 
(i.e., the concatenation of forward and backward hidden states) generated by the BiLSTM.

\subsection{Memory module}

\begin{figure}[!ht]
	\center
    \includegraphics[keepaspectratio=true,scale=0.35]{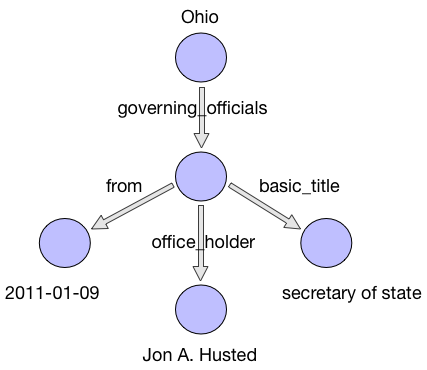}
    \caption{ A
        working example from Freebase. Relations in Freebase have
        hierarchies where high-level ones provide too broad or even
        noisy information about the relation. Thus, we choose to
use the lowest level one.}
  \label{fig:answer_ctx}
\end{figure}

\noindent\textbf{Candidate generation}
Even though all the entities
from the KB could in
principle be candidate answers, this is computationally expensive and
unnecessary in practice. We only consider those
entities which are ``close'' to the main topic entity of a
question.
An answer is the text description (e.g., a name) of an entity node.
For example, \emph{Ohio}
is the topic entity of the question ``Who was the secretary of state of
Ohio in 2011?'' (see \cref{fig:answer_ctx}). After getting the topic entity, 
we collect all the entities connected to it within $h$ hops 
as candidate answers, which we denote as $\{A_i\}_{i=1}^{|A|}$.

\noindent\textbf{KB representation}
For each candidate
answer from the KB, 
we encode three types of information: answer type, path and context.

\textit{Answer type}\quad Entity type information is an important
clue in ranking answers. For example, if a question uses the
interrogative word \emph{where}, then candidate answers with types
relevant to the concept of location are more likely to be correct. 
We use a
BiLSTM to encode its text description to get a $d$-dimensional vector 
$\vec{H}_i^{t_1}$ (i.e., the concatenation of last forward and backward hidden states).

\textit{Answer path}\quad We define an answer path as a sequence
of relations from a candidate answer to a topic entity. 
For example, for the {\em Ohio} question (see \cref{fig:answer_ctx}),
the answer path of $\emph{Jon A. Husted}$ can be either
represented as a sequence of relation ids $[\emph{office\_holder, governing\_officials}]$ or
the text description $[\emph{office, holder, governing, officials}]$. 
We thus encode an answer path as
$\vec{H}_i^{p_1}$ via a BiLSTM, and as $\vec{H}_i^{p_2}$ by computing 
the average of its relation embeddings via a relation embedding layer.

\textit{Answer context}\quad The answer context is defined as the surrounding entities (e.g., sibling nodes) of a
candidate which can help answer questions with constraints.
For example, in \cref{fig:answer_ctx}, the answer
context of $\emph{Jon A. Husted}$ includes the
government position title $\emph{secretary of state}$ and starting
date $\emph{2011-01-09}$. 
However, for simple questions without constraints, the
answer context is unnecessary and can potentially incorporate noise. We tackle this
issue with two strategies: 1) we use a novel {\em importance module} (explained later) to focus on important answer aspects, and 
2) we only consider those context nodes that have overlap with the question. 
Specifically, for each context node (i.e., a sequence of words) of a candidate, 
we first compute the longest common subsequence between it and the question,
we then encode it via a BiLSTM only if we get a non-stopwords substring. 
Finally, the answer context of a candidate answer will be encoded as the
average of all context node representations, which we denote as $\vec{H}_i^c$.

\noindent\textbf{Key-value memory module}
In our model, we use a key-value memory network \cite{miller2016key} to
store candidate answers. Unlike a basic memory network
\cite{weston2014memory}, its addressing stage is based on the key memory 
while the reading stage uses the value memory, which gives greater
flexibility to encode prior knowledge via functionality separation.
Thus, after encoding the answer type, path and context, we apply linear
projections on them as follows:
\begin{align}
    \small
\begin{aligned}
\vec{M}_i^{k_t} & = f_t^k(\vec{H}_i^{t_1}) &
\vec{M}_i^{v_t} & = f_t^v(\vec{H}_i^{t_1}) \\
\vec{M}_i^{k_p} & = f_p^k([\vec{H}_i^{p_1}; \vec{H}_i^{p_2}]) &
\vec{M}_i^{v_p} & = f_p^v([\vec{H}_i^{p_1}; \vec{H}_i^{p_2}])\\
\vec{M}_i^{k_c} & = f_c^k(\vec{H}_i^c) &
\vec{M}_i^{v_c} & = f_c^v(\vec{H}_i^c)
\end{aligned}
\vspace{-0.1in}
\end{align}
where $\vec{M}_i^{k_t}$ and $\vec{M}_i^{v_t}$ are $d$-dimensional key and value
representations of answer type $A_i^t$, respectively.
Similarly, we have key and value representations for
answer path and answer context. 
We denote $\vec{M}$ as a key-value
memory whose row $\vec{M}_i=\{\vec{M}_i^k, \vec{M}_i^v\}$ (both in $\mathbb{R}^{d \times 3}$), where
$\vec{M}_i^k=[\vec{M}_i^{k_t}; \vec{M}_i^{k_p}; \vec{M}_i^{k_c}]$
comprises the keys, and
$\vec{M}_i^v=[\vec{M}_i^{v_t}; \vec{M}_i^{v_p}; \vec{M}_i^{v_c}]$
comprises the values.
Here $[,]$ and $[;]$ denote row-wise
and column-wise concatenations, respectively.

\subsection{Reasoning module}

The reasoning module consists of a generalization module, and our novel {\em two-layered bidirectional 
attention network}
which aims at capturing the two-way interactions between questions and the KB.
The {\em primary attention network} contains the KB-aware
attention module which focuses on the important parts of a
question in light of the KB, and the importance module which
focuses on the important KB aspects in light of the question.
The {\em secondary attention network} 
({\em enhancing module} in \cref{fig:over_arch}) is intended to enhance the question and KB vectors 
by further exploiting the two-way attention.

\begin{figure}[!ht]
        \centering
\includegraphics[keepaspectratio=true,scale=0.3]{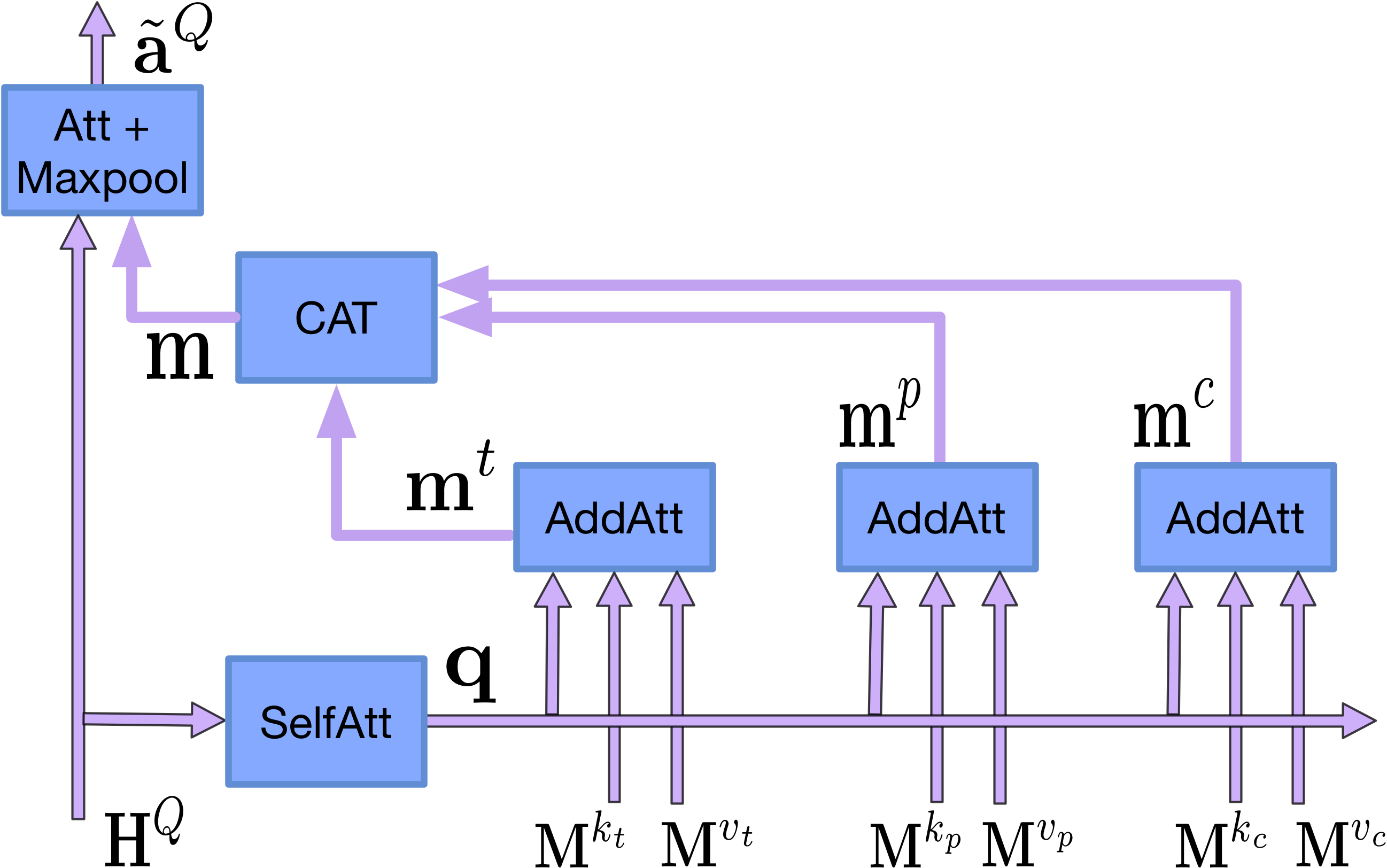}
  \caption{ KB-aware attention module. CAT: concatenation, SelfAtt: self-attention,
      AddAtt: additive attention.}
  \label{fig:kb_aware_att_module}  
\end{figure}


\noindent\textbf{KB-aware attention module}
Not all words in a question are created equal. 
We use a KB-aware attention mechanism to
focus on important components of a question, as shown in
\cref{fig:kb_aware_att_module}. Specifically, we first apply
self-attention (SelfAtt) over all question word vectors
$\vec{H}^Q$ to get a $d$-dimensional question vector $\vec{q}$ as follows
\begin{equation}\label{eq:que_self_att}
\begin{aligned}
\vec{q}  = \BiLSTM([\vec{H}^Q  {\vec{A}^{QQ}}^T, \vec{H}^Q])\\
\vec{A}^{QQ}  = \softmax(({\vec{H}^Q})^T \vec{H}^Q)
\end{aligned}
\end{equation}
where $\softmax$ is applied over the last dimension of an input tensor by default. 
Using question summary $\vec{q}$, we apply another attention (AddAtt) over the memory 
to obtain answer type $\vec{m}_t$, path $\vec{m}_p$ 
and context summary $\vec{m}_c$:
\begin{equation}
\begin{aligned}
\vec{m}_x & = \sum_{i=1}^{|A|}{a_i^x\cdot \vec{M}_i^{v_x}} \\
\vec{a}^x & = \text{Att}_{\text{add}}(\vec{q}, \vec{M}^{k_x})
\end{aligned}
    \label{eq:kb_add_att}
\end{equation}
where $x \in \{t, p, c\}$, and
$\text{Att}_{\text{add}}(\vec{x}, \vec{y}) =
\softmax(\text{tanh}([\vec{x}^T,\vec{y}] \vec{W_{\!1}}) \vec{W_{\!2}})$, with
$\vec{W_{\!1}} \in \mathbb{R}^{2d \times d}$ and $\vec{W_{\!2}} \in \mathbb{R}^{d \times 1}$ being trainable weights.

So far, we have obtained the KB summary $\vec{m}=[\vec{m}_t; \vec{m}_p;
\vec{m}_c]$ in light of the question. We proceed to compute the
question-to-KB attention between question word $q_i$ and KB aspects
as formulated by $\vec{A}^{Qm} = {\vec{H}^Q}^T \vec{m}$. 
By applying max pooling over the last dimension (i.e., the KB aspect
dimension) of $\vec{A}^{Qm}$, that is, $\vec{a}_i^Q = \max_j \vec{A}_{ij}^{Qm}$, 
we select the strongest connection between $q_i$ and the KB. 
The idea behind it is that each word in a question serves a 
specific purpose (i.e., indicating answer type, path or context), and max pooling
can help find out that purpose. We then apply a softmax
over the resulting vector to obtain $\tilde{\vec{a}}^Q$
which is a {\em KB-aware question attention vector} since it
indicates the importance of $q_i$ in light of the KB.


\noindent\textbf{Importance module}
The importance module focuses on important KB aspects 
as measured by their relevance to the questions. 
We start by computing a $|Q|\times |A|\times 3$ attention
tensor $\vec{A}^{QM}$ which indicates the strength of connection between
each pair of $\{q_i, A_j^x\}^{x=\{t,p,c\}}$.
Then, we take the max of the question word dimension
of $\vec{A}^{QM}$ and normalize
it to get an attention matrix $\tilde{\vec{A}}^M$, which
indicates the importance of each answer aspect for each candidate answer. 
After that, we proceed to compute question-aware memory representations
$\tilde{\vec{M}}^k$. Thus, we have:
\begin{align}
\small    
\begin{aligned}
\tilde{\vec{M}}^v & = \{\tilde{\vec{M}}_i^v\}_{i=1}^{|A|} \in
\mathbb{R}^{|A|\times d} &
\tilde{\vec{M}}_i^v & = \sum_{j=1}^3 \vec{M}_{ij}^v\\
\tilde{\vec{M}}^k & = \{\tilde{\vec{M}}_i^k\}_{i=1}^{|A|} \in
\mathbb{R}^{|A|\times d} 	&
\tilde{\vec{M}}_i^k & = \tilde{\vec{A}}_i^M  \vec{M}_i^k \\
\tilde{\vec{A}}^M & = \softmax({\vec{A}^M}^T)^T	&
\vec{A}^M & = \max_i \{{\vec{A}_i^{QM}}\}_{i=1}^{|Q|}	 \\
\vec{A}^{QM} & = \bigl(\vec{M}^k  \vec{H}^Q\bigr)^T\\
\end{aligned}
\end{align}

\noindent\textbf{Enhancing module}
We further enhance the question and KB representations 
by exploiting two-way attention.
We compute the KB-enhanced question representation $\tilde{\vec{q}}$
which incorporates the relevant KB information by applying
max pooling over the last dimension 
(i.e., the answer aspect dimension) 
of $\vec{A}^{QM}$, that is, $\vec{A}_M^Q = \max_k \{\vec{A}_{.,.,k}^{QM}\}_{k=1}^3$, and then normalizing it to get
a question-to-KB attention matrix $\tilde{\vec{A}}_M^Q$ from which we
compute the question-aware KB summary and incorporate it into the
question representation 
$\tilde{\vec{H}}^Q  = \vec{H}^Q + {\tilde{\vec{a}}^Q \odot
(\tilde{\vec{A}}_M^Q  \tilde{\vec{M}}^v)}^T$.
Finally, we obtain a $d$-dimensional KB-enhanced question representation $\tilde{\vec{q}} = \tilde{\vec{H}}^Q  \tilde{\vec{a}}^Q$.

Similarly, 
we compute a question-enhanced KB representation $\overline{\vec{M}}^k$
which incorporates the relevant question information:

\begin{align}
    \begin{aligned}
\overline{\vec{M}}^k & = \tilde{\vec{M}}^k + {\tilde{\vec{a}}^M \odot
            (\tilde{\vec{A}}_Q^M  (\tilde{\vec{H}}^Q)^T)}\\
\tilde{\vec{a}}^M &= (\tilde{\vec{A}}_M^Q)^T  \tilde{\vec{a}}^Q \in
\mathbb{R}^{|A|\times 1} \\
\tilde{\vec{A}}_Q^M & = \softmax({\vec{A}_M^Q}^T) \in \mathbb{R}^{|A|\times
|Q|} 
    \end{aligned}
\end{align}


\noindent\textbf{Generalization module}
We add a one-hop attention process before answering.
We use the question
representation $\tilde{\vec{q}}$ to query over the key memory
$\overline{\vec{M}}^k$ via an attention mechanism, and fetch the most
relevant information from the value memory, which is then
used to update the question vector using a
GRU \cite{cho2014learning}. 
Finally, we apply a residual layer~\cite{he2016deep} (i.e., $y=f(x) + x$) and batch 
normalization (BN)~\cite{ioffe2015batch}, which help the model
performance in practice. Thus, we have
\begin{align}
\begin{aligned}
\hat{\vec{q}} & = \text{BN}(\tilde{\vec{q}} + {\vec{q}}^\prime)	&
{\vec{q}}^\prime & = \text{GRU}(\tilde{\vec{q}}, \tilde{\vec{m}}) \\
\tilde{\vec{m}} & = \sum_{i=1}^{|A|} a_i \cdot \tilde{\vec{M}}_i^v &
\vec{a} & = \text{Att}_{\text{add}}^{\text{GRU}}(\tilde{\vec{q}}, \overline{\vec{M}}^k)
\end{aligned}
\end{align}

\subsection{Answer module}
Given the representation of question $Q$ which is $\hat{\vec{q}}$
and the representation of candidate answers $\{A_i\}_{i=1}^{|A|}$
which is $\{\overline{\vec{M}}_i^k\}_{i=1}^{|A|}$, we compute the matching
score $S(\hat{\vec{q}}, \overline{\vec{M}}_i^k)$ between every pair $(Q,
A_i)$ as $S(\vec{q}, \vec{a}) = \vec{q}^T\cdot \vec{a}$. 
The candidate answers are then ranked by their scores.

\subsection{Training and testing}
\label{sec:training}

\textbf{Training} 
Intermediate modules such as the {\em enhancing module} generate
``premature'' representations of questions (e.g., $\tilde{\vec{q}}$) and 
candidate answers (e.g., $\overline{\vec{M}}^k$).
Even though these intermediate representations are not optimal for answer prediction, we can still use them
along with the final representations to jointly train the model, 
which we find helps the training probably by providing more supervision since we are directly forcing intermediate representations to be helpful for prediction.
Moreover, 
we directly match interrogative words to KB answer types.
A question $Q$ is represented by a 16-dimensional interrogative word (we use
``which'', ``what'', ``who'', ``whose'', ``whom'', ``where'', 
``when'', ``how'', ``why'' and ``whether'') 
embedding $\vec{q}^w$
and a candidate answer $A_i$ is represented by 
entity type embedding $\vec{H}_i^{t_2}$ with 
the same size. We then compute the matching score
$S(\vec{q}^w, \vec{H}_i^{t_2})$ between them.
Although we only have weak labels (e.g., incorrect answers
do not necessarily imply incorrect types) for the type matching task,
and there are no shared representations between two tasks,
we find in practice this strategy helps the training process
as shown in \cref{sec:ablation_study}.

\noindent{\em Loss Function:}
In the training phase, we force positive candidates to
have higher scores than negative candidates
by using a triplet-based loss function:
\begin{equation}
    \begin{aligned}
o = g(\vec{H}^Q \tilde{\vec{a}}^Q, \sum_{j=1}^{3} \vec{M}^k_{.,j})
	+ g(\tilde{\vec{q}}, \overline{\vec{M}}^k)\\
	+ g(\hat{\vec{q}}, \overline{\vec{M}}^k)
	+ g(\vec{q}^w, \vec{H}^{t_2})
	 \end{aligned}
\end{equation}
where \begin{small}
$g(\vec{q}, \vec{M})=\sum_{\substack{a^+ \in A^+\\a^- \in
A^-}}{\ell(S(\vec{q}, \vec{M}_{a^+}), S(\vec{q}, \vec{M}_{a^-}))}$ \end{small},
and
$\ell(y, \hat{y})=\max(0, 1 + \hat{y} - y)$ is a hinge loss function, and
$A^+$ and $A^-$ denote the positive (i.e., correct) and negative (i.e., incorrect) answer
sets, respectively.
Note that at training time, 
the candidate answers are extracted from the KB subgraph of the
gold-standard topic entity,
with the memory size set to
$N_{max}$. 
We adopt the following sampling strategy which works well in practice:
if $N_{max}$ is larger than the number of positive answers $|A^+|$,
we keep all the positive answers and randomly select negative answers to fill up the memory;
otherwise, we randomly select $\min(N_{max} / 2, |A^-|)$ negative answers
and fill up the remaining memory with random positive answers.

\noindent
\textbf{Testing}
At testing time, we need to first find the topic entity. We do this by using the top result returned by a separately trained topic entity predictor (we also compare with the result returned by the Freebase Search API).
Then, the {\em answer module} returns the candidate answer with the highest
scores as predicted answers. Since there can be multiple answers to a given
question, the candidates whose scores are close to the highest score within
a certain margin, $\theta$, are regarded as good answers as well. 
Therefore, we formulate the inference process as follows:
\begin{equation}
\small
\begin{aligned}
\hat{A} = \{\hat{a} \quad | \; \hat{a} \in A\ \&\ \max_{a\prime \in
A}\{S(\hat{\vec{q}}, \overline{\vec{M}}_{a\prime}^k)\} - S(\hat{\vec{q}},
\overline{\vec{M}}_{\hat{a}}^k) < \theta \}
\end{aligned}
\end{equation}
where $\max_{a\prime \in A}\{S(\hat{\vec{q}}, \overline{\vec{M}}_{a\prime}^k)\}$
is the score of the best matched answer and $\hat{A}$ is the predicted answer set.
Note that $\theta$ is a hyper-parameter which controls the degree of tolerance.
Decreasing the value of $\theta$ makes the model become stricter when predicting answers.

\subsection{Topic entity prediction}
\label{sec: topic_ent_pred}

Given a question $Q$, the goal of a topic entity predictor is to find the best topic entity $\hat{c}$ from the candidate set $\{C_i\}_{i=1}^{|C|}$ returned by
external topic entity linking tools (we use the Freebase Search API and S-MART~\cite{yang2016s} in our experiments).
We use a convolutional network (CNN) to encode $Q$ into a $d$-dimensional vector $\vec{e}$.
For candidate topic entity $C_i$, we encode three types of KB aspects, namely, the entity name, entity type and surrounding relations
where both entity name and type are represented as a sequence of words while surrounding relations are represented as a bag of sequences of words.
Specifically, we use three CNNs to encode them into three $d$-dimensional vectors, namely, $\vec{C}_i^n$, $\vec{C}_i^t$ and $\vec{C}_i^{r_1}$.
Note that for surrounding relations, we first encode each of the relations and then compute their average.
Additionally, we compute an average of the relation embeddings via a relation embedding layer which we denote as $\vec{C}_i^{r_2}$.
We then apply linear projections on the above vectors as follows:
\begin{align}
\begin{aligned}
\vec{P}_i^{k} & = f^k([\vec{C}_i^n; \vec{C}_i^t; \vec{C}_i^{r_1}; \vec{C}_i^{r_2}])\\
\vec{P}_i^{v} & = f^v([\vec{C}_i^n; \vec{C}_i^t; \vec{C}_i^{r_1}; \vec{C}_i^{r_2}])\\
\end{aligned}
\end{align}
where $\vec{P}_i^{k}$ and $\vec{P}_i^{v}$ are $d$-dimensional key and value
representations of candidate $C_i$, respectively.
Furthermore, we compute the updated question vector $\hat{\vec{e}}$ using the generalization module mentioned earlier.
Next, we use a dot product to compute the similarity score between $Q$ and $C_i$. 
A triplet-based loss function is used as formulated by
$o = g(\vec{e}, \vec{P}_i^{k})
	+ g(\hat{\vec{e}}, \vec{P}_i^{k})$
where $g(.)$ is the aforementioned hinge loss function.
When training the predictor, along with the candidates returned from external entity linking tools, we do negative sampling (using string matching) to get more supervision.
In the testing phase, the candidate with the highest score is returned as the best topic entity and no negative sampling is applied.

\section{Experiments}

This section provides an extensive evaluation of our proposed BAMnet model
against state-of-the-art KBQA methods.
The implementation of BAMnet is available at \url{https://github.com/hugochan/BAMnet}.

\subsection{Data and metrics}
We use the Freebase KB and the WebQuestions dataset, described below:

\noindent\textit{Freebase} \quad This is a large-scale KB
\cite{freebase:datadumps} that
consists of general facts organized as subject-property-object triples.
It has 41M non-numeric entities, 19K properties, and 596M assertions.

\noindent\textit{WebQuestions} \quad This dataset \cite{berant2013semantic}
(\url{nlp.stanford.edu/software/sempre}) contains 3,778 training
examples and 2,032 test examples. We further split the training
instances into a training set and development set via a 80\%/20\% split.
Approximately 85\% of questions can be directly
answered via a single FreeBase predicate. Also, each question can have
multiple answers.
In our experiments, we use a development version of the dataset
\cite{baudis2016},
which additionally provides (potentially noisy) entity mentions for each question.

Following \cite{berant2013semantic}, macro F1 scores (i.e., the average of F1 scores over all questions) are reported
on the WebQuestions test set.

\subsection{Model settings}

%
When constructing the vocabularies of words, entity types or relation
types, we only consider those questions and their corresponding KB
subgraphs appearing in the training and validation sets. The vocabulary
size of words is $V=100,797$. There are 1,712 entity types and 4,996
relation types in the KB subgraphs. Notably, in FreeBase, one entity
might have multiple entity types. We only use the first one available,
which is typically the most concrete one. For those non-entity nodes
which are boolean values or numbers, we use ``bool'' or ``num'' as their
types, respectively. 

We also adopt a \textbf{query delexicalization} strategy where 
for each question, the topic entity mention as well as constraint entity mentions 
(i.e., those belonging to ``date'', ``ordinal'' or ``number'')
are replaced with their types.
When encoding KB context, if the overlap belongs to the above types, 
we also do this delexicalization, which will guarantee it matches up with
the delexicalized question well in the embedding space.

Given a topic entity, we extract its 2-hop subgraph
(i.e., $h=2$) to collect candidate answers, which is sufficient for \textit{WebQuestions}. At training time, the memory
size is limited to $N_{max} = 96$ candidate answers (for the sake of efficiency). If
there are more potential candidates, we do random sampling as mentioned
earlier. We initialize word embeddings with pre-trained GloVe
vectors~\cite{pennington2014glove}
with word embedding size $d_v = 300$.
The relation embedding size $d_p$, entity type embedding size $d_t$ 
and hidden size $d$ are set as 128, 16 and 128, respectively.
The dropout rates on the word embedding layer, question encoder side and the answer encoder
side are 0.3, 0.3 and 0.2, respectively. The batch size is set as 32, and answer
module threshold $\theta=0.7$. 
As for the topic entity prediction, we use the same hyperparameters.
For each question, there are 15 candidates after negative sampling in the training time.
When encoding a question, we use a CNN with filter sizes 2 and 3.
A linear projection is applied to merge features extracted with different filters.
When encoding a candidate aspect, we use a CNN with filter size 3.
Linear activation and max-pooling are used together with CNNs.
In the training process, we use the Adam optimizer
\cite{kingma2014adam} to train the model. The initial learning rate is
set as 0.001 which is reduced by a factor of 10 if no improvement is
observed on the validation set in 3 consecutive epochs. The training
procedure stops if no improvement is observed on the validation set in
10 consecutive epochs. 
The hyper-parameters are tuned on the development set.

\subsection{Performance comparison}
As shown in \cref{table:webq_results},
our method can achieve an F1 score of 0.557 when the gold topic
entity is known, which gives an upper bound of our model performance. 
When the gold topic entity is unknown, 
we report the results using:
1) the Freebase Search API, which achieves a recall@1
score of 0.857 on the test set for topic entity linking,
and 2) the topic entity predictor, which achieves a recall@1
score of 0.898 for entity retrieval.

As for the performance of BAMnet on WebQuestions, it 
achieves an F1 score of 0.518 using the topic entity predictor, which is significantly better than the F1 score of 0.497 using the Freebase Search API.
We can observe that BAMnet significantly outperforms previous
state-of-the-art IR-based methods, which conclusively demonstrates the effectiveness
of modeling bidirectional interactions between questions and the KB.

\begin{table}[!ht]
  \centering
  \begin{tabular}{|c|c|}
    \toprule
     Methods (ref) & Macro $F_1$  \\
    \hline
    \multicolumn{2}{|c|}{SP-based}\\
    \midrule 
    \cite{berant2013semantic}  & 0.357    \\
    \cite{yao2014information} & 0.443      \\
    \cite{wang2014overview}       & 0.453  \\
    \cite{bast2015more} & 0.494 \\
    \cite{berant2015imitation} & 0.497 \\
    \cite{yih2015semantic} & {\bf 0.525}\\
    \cite{reddy2016transforming} & 0.503 \\
    \cite{yavuz2016improving} & 0.516\\
    \cite{bao2016constraint} & 0.524 \\
    \cite{feng2016hybrid} & 0.471 \\
    \cite{reddy2017universal} & 0.495 \\
    \cite{abujabal2017automated} & 0.510 \\
   \cite{hu2018answering} & 0.496 \\
    \hline
    \multicolumn{2}{|c|}{IR-based}\\
	 \midrule 
     \cite{bordes2014question} & 0.392 \\
     \cite{yang2014joint} & 0.413 \\
     \cite{dong2015question} & 0.408 \\ %
     \cite{bordes2015large} & 0.422 \\
     \cite{xu2016question} & {\bf 0.471} \\
    \cite{hao2017end} & 0.429 \\
    \hline
    \multicolumn{2}{|c|}{Our Method: BAMnet}\\
    \midrule %
    w/ gold topic entity & 0.557 \\
	w/ Freebase Search API & 0.497 \\
    w/ topic entity predictor& {\bf 0.518} \\
    \bottomrule
  \end{tabular}
   \caption{ Results on the WebQuestions test set. Bold: best
  in-category performance.}
  \label{table:webq_results}
\end{table} 

It is important to note that unlike the state-of-the-art SP-based methods,
BAMnet relies on no external resources and very few hand-crafted features, 
but still remains competitive with those approaches.
Based on careful hand-drafted rules, some
SP-based methods \cite{bao2016constraint,yih2015semantic} can
better model questions with constraints and
aggregations. For example, \cite{yih2015semantic} applies many
manually designed rules and features to improve performance on
questions with constraints and aggregations, and \cite{bao2016constraint}
directly models temporal
(e.g., ``after 2000''), ordinal (e.g., ``first'') and
aggregation constraints (e.g., ``how many'')
by adding detected constraint nodes to query graphs.
In contrast, our method is end-to-end, with very few hand-crafted rules.

Additionally, \cite{yavuz2016improving,bao2016constraint} 
train their models on external Q\&A datasets to get extra supervision.  
For a fairer comparison, we only show their results
without training on external Q\&A datasets. 
Similarly, for hyhrid systems \cite{feng2016hybrid,xu2016question}, 
we only report results without using Wikipedia free text.
It is interesting to note that both 
\cite{yih2015semantic} and \cite{bao2016constraint} also use the 
ClueWeb dataset for learning more accurate semantics. The F1 score of 
\cite{yih2015semantic} drops from 0.525 to 0.509 if ClueWeb information is removed. To summarize, BAMnet achieves state-of-the-art performance of 0.518 without recourse to any external resources and relies only on very few hand-crafted features. If we assume gold-topic entities are given then BAMnet achieves an F1 of 0.557.

\begin{table}[!ht]
  \centering
  \begin{tabular}{ll}
    \toprule
    Methods      & Macro $F_1$  \\
    \midrule %
    all & \textbf{0.557}\\
    w/o two-layered bidirectional attn  & 0.534 \\
	w/o kb-aware attn (+self-attn) & 0.544 \\
	w/o importance module & 0.540 \\
	w/o enhancing module& 0.550\\
	w/o generalization module & 0.542\\
	w/o joint type matching & 0.545\\
	w/o topic entity delexicalization & 0.529\\
	w/o constraint delexicalization & 0.554\\
    \bottomrule
  \end{tabular}
     \captionof{table}{ Ablation results on the WebQuestions test set.
  Gold topic entity is assumed to be known.}
  \label{table:ablation_results}
\end{table}

\subsection{Ablation study} \label{sec:ablation_study}
We now discuss the performance impact of the different modules and strategies in BAMnet.
Note that gold topic entity is assumed to be known when we do this ablation study, 
because the error introduced by topic entity prediction might reduce the 
real performance impact of a module or strategy.
As shown in \cref{table:ablation_results}, significant performance drops were observed after
turning off some key attention modules, which confirms that the real power of our method comes from 
the idea of hierarchical two-way attention. 
As we can see, when turning off the {\em two-layered bidirectional attention network}, the model performance
drops from 0.557 to 0.534.
Among all submodules in the attention network, the {\em importance module} is the
most significant since the F1 score drops to 0.540 without it, thereby confirming the effectiveness of modeling the
query-to-KB attention flow. On the flip side, 
the importance of modeling the KB-to-query attention flow is confirmed
by the fact that replacing the {\em KB-aware attention module} with self-attention significantly degrades the performance.
Besides, the secondary attention layer, the {\em enhancing module}, also contributes to the overall model performance.
Finally, we find that the topic entity delexicalization strategy has a big influence on the model performance while the
constraint delexicalization strategy only marginally boosts the performance.

\begin{table*}[!h]
        \small
 \centering
 \renewcommand{\arraystretch}{1.}	
  \begin{tabular}{|p{1.2cm}|c|c|c|c|} \hline
       \toprule
    \makecell{KB\\ Aspects} & Questions & BAMnet w/o BiAttn. & BAMnet & Gold Answers  \\
    \hline
    \multirow{2}{.1cm}{\makecell{Answer \\Type}}
    & \makecell{What degrees did Obama \\get in college?} & \makecell{Harvard Law School, \\Columbia University, \\Occidental College} & \makecell{Bachelor of Arts, \\Juris Doctor, \\Political Science}  & \makecell{Juris Doctor,\\ Bachelor of Arts} \\ \cline{2-5}
    & \makecell{What music period did \\Beethoven live in?} & \makecell{Austrian Empire, \\Germany, Bonn}& \makecell{Classical music, \\Opera} &  \makecell{Opera, \\Classical music} \\ 
    	\hline
    
     \multirow{2}{.1cm}{\makecell{Answer \\Path}}
     & \makecell{Where did Queensland \\get its name from?} & \makecell{Australia} & \makecell{Queen Victoria}  & \makecell{Queen Victoria} \\ \cline{2-5}
          & \makecell{Where does Delaware \\river start?} & \makecell{Delaware Bay} & \makecell{West Branch\\ Delaware River,\\ Mount Jefferson} & \makecell{West Branch\\ Delaware River,\\ Mount Jefferson}
          \\ \hline
    	 
      \multirow{2}{.1cm}{\makecell{Answer \\Context}}
     & \makecell{What are the major \\cities in Ukraine?} & \makecell{Kiev, Olyka,\\ ...\\ Vynohradiv, Husiatyn}& \makecell{Kiev}& \makecell{Kiev}  \\ \cline{2-5}
    	& \makecell{Who is running for vice president\\ with Barack Obama 2012?} 
    	 & \makecell{David Petraeus}
    	 & \makecell{Joe Biden} & \makecell{Joe Biden} \\ \hline
  \end{tabular}
        \caption{Predicted answers of BAMnet w/ and w/o bidirectional attention on the WebQuestions test set.}
        \label{table:comparison_of_examples}
\end{table*}

\begin{figure}[!h]
  \centering
  \includegraphics[keepaspectratio=true,scale=0.26]{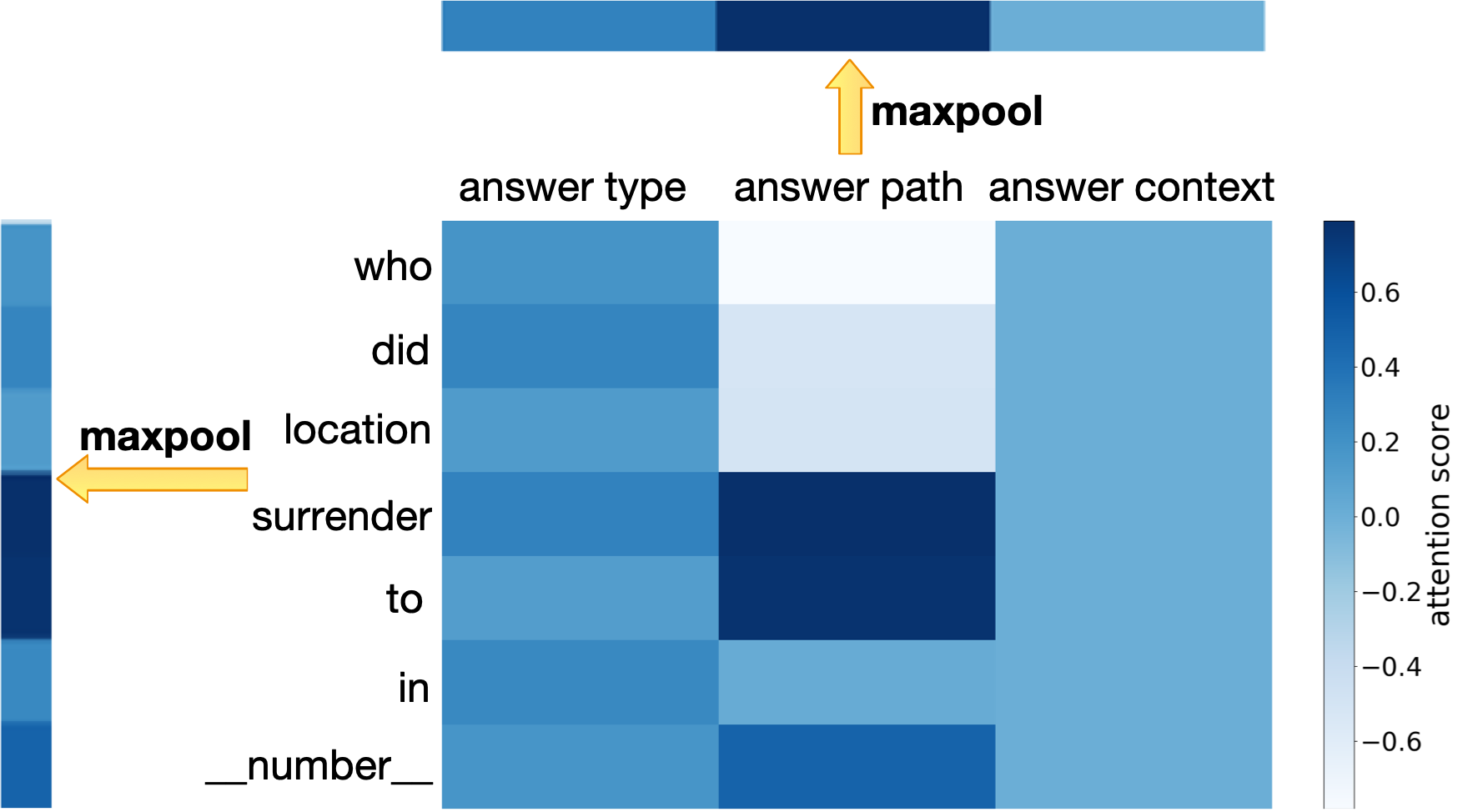}
  \caption{ Attention heatmap generated by the reasoning module. Best viewed in color.}
  \label{fig:att_heatmap}
\end{figure}

\subsection{Interpretability analysis}
Here, we show that our method does capture the mutual interactions between question
words and KB aspects, by visualizing the attention matrix $\vec{A}^{QM}$ produced by
the {\em reasoning module}. \cref{fig:att_heatmap} shows the attention heatmap
generated for a test question ``who did location surrender to in \_\_number\_\_''
(where ``location'' and ``\_\_number\_\_'' are entity types which replace the topic entity
mention ``France'' and  the constraint entity mention ``ww2'', respectively in the original question). 
As we can see, the attention network successfully detects
the interactions between ``who'' and answer type, ``surrender to'' and answer path, and focuses more on those words when encoding the question.
%

To further examine the importance of the two-way flow of interactions, in \cref{table:comparison_of_examples}, we show the predicted answers of BAMnet with and without the {\em two-layered bidirectional attention network} on samples questions from the WebQuestions test set.
We divide the questions into three categories based on which kind of KB aspect is the most crucial for answering them. As we can see, compared to the simplified version which is not equipped with bidirectional attention, our model is more capable of answering all the three types of questions.

\subsection{Error analysis}
To better examine the limitations of our
approach, we randomly sampled 100 questions on which our method
performed poorly (i.e., with per-question F1 score less than 0.6), and
categorized the errors. 
We found that around 33\% of errors are due to
label issues of gold answers and are not real mistakes. 
This includes incomplete and erroneous labels, and also alternative correct answers. 
Constraints are another source of errors (11\%), with temporal
constraints accounting for most. Some questions have implicit temporal (e.g., tense)
constraints which our method does not model. 
A third source of error is what we term type errors (13\%), for which our method generates
more answers than needed because of poorly utilizing answer type information. 
Lexical gap is another source of errors (5\%). 
Finally, other sources of errors (38\%) include topic entity prediction error, question ambiguity, incomplete answers and other
miscellaneous errors.

\section{Conclusions and future work}
We introduced a novel and effective bidirectional
attentive memory network for the purpose of KBQA. To our best knowledge,
we are the first to model the mutual interactions between questions and
a KB, which allows us to distill
the information that is the most relevant to answering the questions on
both sides of the question and KB. Experimental results show that our method
significantly outperforms previous IR-based methods while remaining
competitive with hand-crafted SP-based methods. 
Both ablation
study and interpretability analysis verify the effectiveness of the idea of
modeling mutual interactions. In addition, our error analysis shows that
our method actually performs better than what the evaluation metrics
indicate. 

In the future, we would like to explore effective ways of
modeling more complex types of constraints (e.g., ordinal, comparison and aggregation).

\section*{Acknowledgments}
This work is supported by IBM Research AI through the IBM AI Horizons Network. We thank the anonymous reviewers for their constructive suggestions.

\bibliography{naaclhlt2019}
\bibliographystyle{acl_natbib}
\end{document}